\documentclass[letterpaper, 10 pt, conference]{ieeeconf}  

\IEEEoverridecommandlockouts                              

\usepackage[english]{babel}
\usepackage{graphics}
\usepackage{pgfplots}
\usepackage{pgfplotstable}
\usepackage{tikz}
\usepackage[cmex10]{amsmath}
\usepackage{amsfonts}
\usepackage{bm}
\usepackage{xfrac}
\usepackage{float}
\usepackage[hidelinks]{hyperref} 
\usepackage{booktabs}    
\usepackage[separate-uncertainty=true,detect-mode=true]{siunitx}     
\usepackage{color}
\usepackage{subcaption}
\usepackage[capitalize]{cleveref}
\usepackage[backend=bibtex,style=ieee,sortcites]{biblatex}
\usepackage{algorithm}
\usepackage[noend]{algpseudocode}
\usepackage{mathtools}
\usepackage{amssymb}
\usepackage{amsmath}

\usepackage{graphicx}

\usetikzlibrary{arrows}
\usetikzlibrary{fit}
\usetikzlibrary{matrix}
\usetikzlibrary{positioning}
\usetikzlibrary{intersections}
\usetikzlibrary{graphs}
\usetikzlibrary{calc}
\usetikzlibrary{patterns}
\usetikzlibrary{decorations.pathreplacing}
\usepgfplotslibrary{groupplots}

\pgfplotsset{/pgfplots/ybar legend/.style={
    /pgfplots/legend image code/.code={%
       \draw[##1,/tikz/.cd,yshift=-0.25em]
        (0cm,0cm) rectangle (3pt,0.8em);},
   },
   ylabsh/.style={ 
    every axis y label/.style={
      at={(0,0.5)}, xshift=#1, rotate=90
      }
  },
}

\definecolor{colorBluish}{RGB}{101,161,216}
\definecolor{colorBluePastel}{RGB}{139,185,226}
\definecolor{colorGreenish}{RGB}{106,130,94}
\definecolor{colorGreenPastel}{RGB}{150,194,154}
\definecolor{colorRedish}{RGB}{140,21,21}
\definecolor{colorRedPastel}{RGB}{164,117,129}
\definecolor{colorOrangish}{RGB}{237,125,49}
\definecolor{colorOrangePastel}{RGB}{213,169,143}
\definecolor{colorPurplePastel}{RGB}{216,191,216}

\definecolor{colorA}{HTML}{AAAAAA}
\definecolor{colorB}{HTML}{999999}
\definecolor{colorC}{HTML}{777777}
\definecolor{colorD}{HTML}{555555}
\definecolor{colorE}{HTML}{444444}
\definecolor{colorF}{HTML}{222222}
\definecolor{colorG}{HTML}{000000}
\definecolor{colorH}{HTML}{999999}
\definecolor{colorI}{HTML}{777777}

\newenvironment{texthook}[1]{}{}

\DeclareMathOperator*{\E}{\mathbb{E}}

\newcommand{\paren}[1]{\mathopen{}\mathclose\bgroup\left(#1\aftergroup\egroup\right)}
\newcommand{\brock}[1]{\mathopen{}\mathclose\bgroup\left[#1\aftergroup\egroup\right]}
\newcommand{\curly}[1]{\mathopen{}\mathclose\bgroup\left\{#1\aftergroup\egroup\right\}}

\AtEveryBibitem{
  \clearfield{doi}
  \clearfield{issn}
  \ifentrytype{inproceedings}{
    \clearlist{address}
    \clearlist{publisher}
    \clearname{editor}
    \clearlist{organization}
    \clearfield{url}
    \clearfield{pages}
    \clearlist{location}
  }{}
}

\renewbibmacro*{bbx:savehash}{}
\renewbibmacro{finentry}{%
  \iffieldequalstr{entrykey}{liu2014latpos}
   {\finentry\newpage}
   {\finentry}
}

\newsavebox\myboxA
\newsavebox\myboxB
\newlength\mylenA
\newcommand*\xoverline[2][0.75]{%
    \sbox{\myboxA}{$\m@th#2$}%
    \setbox\myboxB\null
    \ht\myboxB=\ht\myboxA%
    \dp\myboxB=\dp\myboxA%
    \wd\myboxB=#1\wd\myboxA
    \sbox\myboxB{$\m@th\overline{\copy\myboxB}$}
    \setlength\mylenA{\the\wd\myboxA}
    \addtolength\mylenA{-\the\wd\myboxB}%
    \ifdim\wd\myboxB<\wd\myboxA%
       \rlap{\hskip 0.5\mylenA\usebox\myboxB}{\usebox\myboxA}%
    \else
        \hskip -0.5\mylenA\rlap{\usebox\myboxA}{\hskip 0.5\mylenA\usebox\myboxB}%
    \fi}
\makeatother

\AtEveryBibitem{
	\clearlist{address}
	\clearlist{organization}
	\clearfield{url}
	\clearfield{doi}
	\clearlist{location}
	\clearlist{issn}
	\clearfield{month}
}

\title{\LARGE \bf
Imitating Driver Behavior with Generative Adversarial Networks
}

\author{Alex Kuefler$^{1}$, Jeremy Morton$^{2}$, Tim Wheeler$^{2}$, and Mykel Kochenderfer$^{2}$
\thanks{$^{1}$Alex Kuefler is in the Symbolic Systems Program at Stanford University, Stanford, CA 94305, USA
        {\tt\small akuefler@stanford.edu}}%
\thanks{$^{2}$Are in the department of Aeronautics and Astronautics at Stanford University, Stanford, CA 94305, USA
        {\tt\small \{jmorton2, wheelert, mykel\}@stanford.edu}}%
}

\begin{document}

\maketitle
\thispagestyle{empty}
\pagestyle{empty}

\begin{abstract}
The ability to accurately predict and simulate human driving behavior is critical for the development of intelligent transportation systems.
Traditional modeling methods have employed simple parametric models and behavioral cloning.
This paper adopts a method for overcoming the problem of cascading errors inherent in prior approaches, resulting in realistic behavior that is robust to trajectory perturbations.
We extend Generative Adversarial Imitation Learning to the training of recurrent policies, and we demonstrate that our model outperforms rule-based controllers
and maximum likelihood models in realistic highway simulations. Our model both reproduces emergent behavior of human drivers, such as lane change rate, while maintaining realistic control over long time horizons.
\end{abstract}

\section{INTRODUCTION}

Accurate human driver models are critical for realistic simulation of driving scenarios, and have the potential to significantly advance research in automotive safety.
Traditionally, human driver modeling has been the subject of both rule-based and data-driven approaches.
Early rule-based attempts include parametric models of car following behavior, with strong built-in assumptions about road conditions~\cite{seddon1972program} or driver behavior~\cite{gipps1981behavioural}.
The Intelligent Driver Model (IDM)~\cite{treiber2000congested} extended this work by capturing asymmetries between acceleration and deceleration, preferred free road and bumper-to-bumper headways, and realistic braking behavior. 
Such car-following models were later extended to multilane conditions with controllers like MOBIL~\cite{kesting2007general}, which maintains a utility function and ``politeness parameter" to capture intelligent driver behavior in both acceleration and turning.
These controllers are all largely characterized by smooth, collision-free driving, but rely on assumptions about driver behavior and a small set of parameters that may not generalize well to diverse driving scenarios.

In contrast, imitation learning (IL) approaches rely on data typically provided through human demonstration in order to learn a policy that behaves similarly to an expert. These policies can be represented with expressive models, such as neural networks, with less interpretable parameters than those used by rule-based methods. Prior human behavior models for highway driving have relied on behavioral cloning (BC), which treats IL as a supervised learning problem, fitting a model to a fixed dataset of expert state-action pairs ~\cite{wheeler2016propagation, Lefevre2014, gindele2013learning, agamennoni2012}.
ALVINN~\cite{pomerleau1989alvinn}, an early BC approach, trained a neural network to map raw images and rangefinder inputs to discrete turning actions.
Recent advances in computing and deep learning have allowed this approach to scale to realistic scenarios, such as parking lot, highway, and markerless road conditions~\cite{bojarski2016end}.
These BC approaches are conceptually sound~\cite{syed2007game}, but tend to fail in practice as small inaccuracies compound during simulation.
Inaccuracies lead the policy to states that are underrepresented in the training data (e.g., an ego-vehicle edging towards the side of the road), which leads to yet poorer predictions, and ultimately to invalid or unseen situations (e.g., off-road driving).
This problem of cascading errors~\cite{ross2010efficient} is well-known in the sequential decision making literature and has motivated work on alternative IL methods, such as inverse reinforcement learning (IRL)~\cite{abbeel2004apprenticeship}.

Inverse reinforcement learning assumes that the expert follows an optimal policy with respect to an unknown reward function.
If the reward function is recovered, one can simply use RL to find a policy that behaves identically to the expert.
This imitation extends to unseen states; in highway driving a vehicle that is perturbed toward the lane boundaries should know to return toward the lane center.
IRL thus generalizes much more effectively and does not suffer from many of the problems of BC.
Because of these benefits, some recent efforts in human driver modeling emphasize IRL~\cite{gonzalez2016high, sadigh2016planning}.
However, IRL approaches are typically computationally expensive in their recovery of an expert cost function.
Instead, recent work has attempted to imitate expert behavior through direct policy optimization, without first learning a cost function~\cite{ho2016model, ho2016generative}. Generative Adversarial Imitation Learning (GAIL)~\cite{ho2016generative} in particular has performed well on a number of benchmark tasks, leveraging the insight that expert behavior can be imitated by training a policy to produce actions that a binary classier mistakes for those of an expert.

In this work, we apply GAIL to the task of modeling human highway driving behavior. Our major contributions are twofold. First, we extend GAIL to the optimization of recurrent neural networks, showing that such policies perform with greater fidelity to expert behavior than feedforward counterparts. Second, we apply our models to a realistic highway simulator, where expert demonstrations are given by real-world driver trajectories included in the NGSIM dataset~\cite{systematics2006ngsim, systematics2007ngsim}. We demonstrate that policy networks optimized by GAIL capture many desirable properties of earlier IL models, such as reproducing emergent driver behavior and assigning high likelihood to expert actions, while simultaneously reducing collision and off-road rates necessary for long horizon highway simulations. Unlike past work, our model learns to map raw LIDAR readings and simple, hand-picked road features to continuous actions, adjusting only turn-rate and acceleration each time step.

\section{PROBLEM FORMULATION}

We regard highway driving as a sequential decision making task in which the driver obeys a stochastic policy $\pi(a \mid s)$ mapping observed road conditions $s$ to a distribution over driving actions $a$.
Given a class of policies $\pi_\theta$ parameterized by $\theta$, we seek to find the policy that best mimics human driving behavior.
We adopt an IL approach to infer this policy from a dataset consisting of a sequence of state-action tuples $(s_t, a_t)$.
IL can be performed using BC or reinforcement learning.

\subsection{Behavioral Cloning}
Behavioral cloning solves a regression problem in which the policy parameterization is obtained by maximizing the likelihood of the actions taken in the training data.
BC works well for states adequately covered by the training data.
It is forced to generalize when predicting actions for states with little or no data coverage, which can lead to poor behavior.
Unfortunately, even if simulations are initialized in common states, the stochastic nature of the policies allow small errors in action predictions to compound over time, eventually leading to states that human drivers infrequently visit and are not adequately covered by the training data.
Poorer predictions can cause a feedback cycle known as cascading errors~\cite{bagnell2015invitation}.

In a highway driving context, cascading errors can lead to off-road driving and collisions.  Datasets rarely contain information about how human drivers behave in these situations, which can lead BC policies to act erratically when they encounter such states.

Behavioral cloning has been successfully used to produce driving policies for simple behaviors such as car-following on freeways, in which the state and action space can be adequately covered by the training set.
When applied to learning general driving models with nuanced behavior and the potential to drive out of lane, BC only produces accurate predictions up to a few seconds~\cite{Lefevre2014,wheeler2016propagation}.

\subsection{Reinforcement Learning}
Reinforcement learning (RL) instead assumes that drivers in the real world follow an expert policy $\pi_{E}$ whose actions maximize the expected, global return

\begin{equation}
\label{eq:expected_reward}
R(\pi,r) = \mathbb{E}_{\pi}\brock{\sum_{t=0}^{T}\gamma^t r(s_t, a_t)}
\end{equation}

\noindent
weighted by a discount factor $\gamma \in [0,1)$.
The local reward function $r(s_t, a_t)$ may be unknown, but fully characterizes expert behavior such that any policy optimizing $R(\pi,r)$ will perform indistinguishably from $\pi_{E}$.

Learning with respect to $R(\pi,r)$ has several advantages over maximum likelihood BC in the context of sequential decision making~\cite{sutton1998reinforcement}. First, $r(s_t,a_t)$ is defined for all state-action pairs, allowing an agent to receive a learning signal even from unusual states. In contrast, BC only receives a learning signal for those states represented in a labeled, finite dataset. Second, unlike labels, rewards allow a learner to establish preferences between mildly undesirable behavior (e.g., hard braking) and extremely undesirable behavioral (e.g., collisions). And finally, RL maximizes the global, expected return on a trajectory, rather than local instructions for each observation. Once preferences are learned, a policy may take mildly undesirable actions now in order to avoid awful situations later. As such, reinforcement learning algorithms provide robustness against cascading errors.

\section{POLICY REPRESENTATION}

Our learned policy must be able to capture human driving behavior, which involves:

\begin{itemize}
  \item \textit{Non-linearity} in the desired mapping from states to actions (e.g., large corrections in steering to avoid collisions caused by small changes in the current state).
  \item \textit{High-dimensionality} of the state representation, which must describe properties of the ego-vehicle, in addition to surrounding cars and road conditions.
  \item \textit{Stochasticity} because humans may take different actions each time they encounter a given traffic scene.
\end{itemize}

\begin{figure*}
\centering
    \begin{minipage}[t]{3in}
      \centering
\includegraphics[width=7cm]{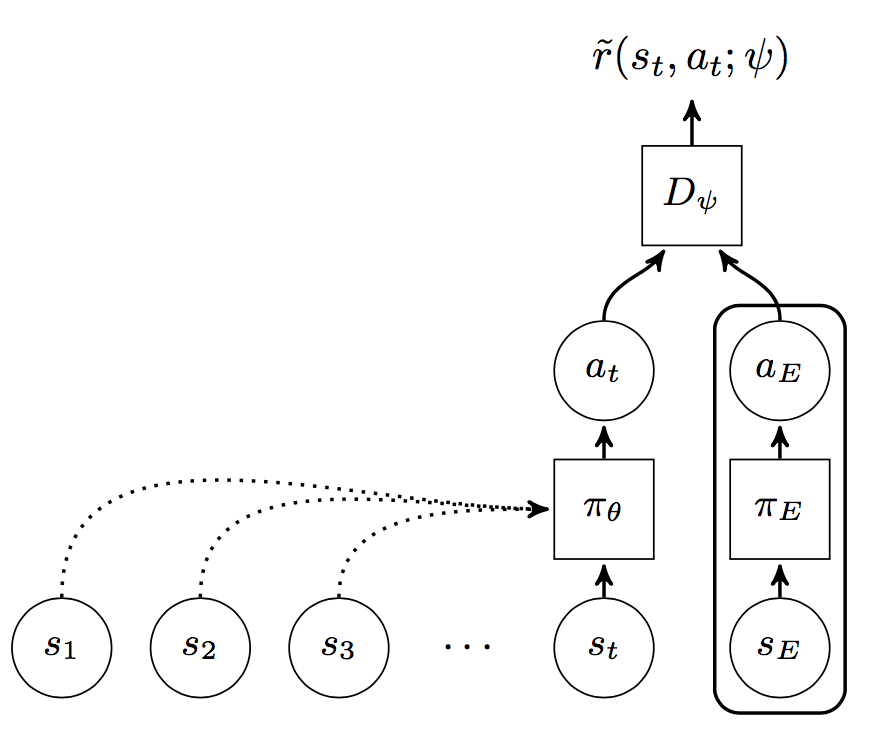}
       \caption{
         GAIL diagram. Circles represent values, rectangles represent operations, and dotted arrows represent recurrence relations.
         Whereas policy $\pi_\theta$ states $s_t$ and actions $a_t$ are evaluated by discriminator $D_\psi$ during both training of $\pi_\theta$ and $D_\psi$, expert pairs (boxed) are sampled only while training $D_\psi$.}
    \label{gail_schematic}
    \end{minipage}
   \quad
    \begin{minipage}[t]{3in}
      \centering
\includegraphics[width=4cm]{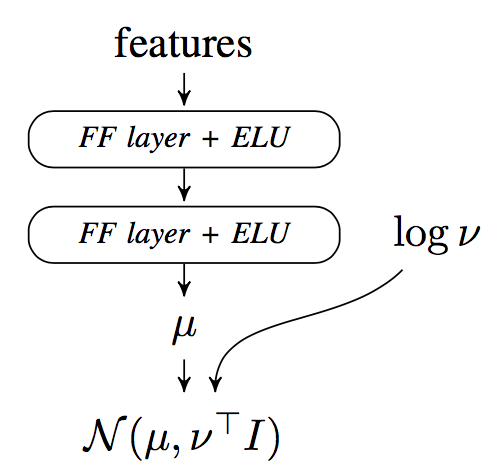}
      \caption{Architecture for the feedforward multilayer perceptron driving policy.  The network output $\mu$ and covariance parameters $\nu$ are used to construct a Gaussian distribution over driver actions.}
      \label{fig:MLPstructure}
    \end{minipage}
\end{figure*}

To address the first and second points, we represent all learned policies $\pi_{\theta}$ using neural networks. To address the third  point, we interpret the network's real-valued outputs given input $s_t$ as the mean $\mu_t$ and logarithm of the diagonal covariance $\log \nu_t$ of a Gaussian distribution.
Actions are chosen by sampling $a_t \sim \pi_{\theta}(a_t \mid s_t)$.
An example feedforward model is shown in~\cref{fig:MLPstructure}.
We evaluate both feedforward and recurrent network architectures.

Feedforward neural networks directly map inputs to outputs.
The most common architecture, multilayer perceptrons (MLPs), consist of alternating layers of tunable weights and element-wise nonlinearities.
Neural networks have gained widespread popularity due to their ability to learn robust hierarchical features from complicated inputs~\cite{lee2009convolutional, krizhevsky2012imagenet}, and have been used in automotive behavioral modeling for action prediction in car-following contexts~\cite{hongfei2003develop, panwai2007neural, khodayari2012modified, Lefevre2014, morton2016analysis}, lateral position prediction~\cite{liu2014latpos}, and maneuver classification~\cite{boyraz2007signal}.

The feedforward MLP is limited in its ability to adequately address partially observable environments.
In real world driving, sensor error and occlusions may prevent the driver from seeing all relevant parts of the driving state.
By maintaining sufficient statistics of past observations in memory, recurrent policies~\cite{wierstra2010recurrent, heess2015memory} disambiguate perceptually similar states by acting with respect to histories of, rather than individual, observations.
In this work, we represent recurrent policies using Gated Recurrent Unit (GRU) networks due to their comparable performance with fewer parameters than other architectures~\cite{chung2014empirical}.

We use similar architectures for the feedforward and recurrent policies. The recurrent policies consist of five feedforward layers that decrease in size from 256 to 32 neurons, with an additional GRU layer consisting of 32 neurons. Exponential linear units (ELU) were used throughout the network, which have been shown to combat the vanishing gradient problem while supporting a zero-centered distribution of activation vectors ~\cite{clevert2015fast}.
  The MLP policies have the same architecture, except the GRU layer is replaced with an additional feedforward layer.  For each network architecture, one policy is trained through BC and one policy is trained through GAIL.  In all, we trained four neural network policies: GAIL GRU, GAIL MLP, BC GRU, and BC MLP.

\section{POLICY OPTIMIZATION}

Contrary to BC, which is trained with traditional regression techniques, reinforcement learning policies do not have training labels for individual actions.
Controller performance is instead evaluated by expected return.
This approach is problematic in modeling human drivers, as the reward function $r(s_{t},a_{t})$ is unknown.
We first discuss a method for training a policy with a known reward function and then provide a method for learning the reward function.

\subsection{Trust Region Policy Optimization}

Policy gradient algorithms are a particularly effective class of reinforcement learning techniques for optimizing differentiable policies, including neural networks.
As with standard backpropagation, network parameters are optimized using gradient-based updates, but the gradient can only be approximated using simulated rollouts of the policy interacting with the environment.

This empirical gradient estimate typically exhibits a high amount of variance.
In practice, this variance can cause parameter updates that do not improve or even reduce performance.
In this work, we use Trust Region Policy Optimization (TRPO) to learn our human driving policies~\cite{schulman2015trust}.
TRPO updates policy parameters through a constrained optimization procedure that enforces that a policy cannot change too much in a single update, and hence limits the damage that can be caused by noisy gradient estimates.

Although the true reward function that governs the behavior of any particular human driver is unknown, domain knowledge can be used to craft a surrogate reward function such that a policy maximizing this quantity will realize a similar stochastic state-action mapping as $\pi_{E}$.
Drivers avoid collisions and going off road, while also favoring smooth driving and minimizing lane-offset.
If such features can be combined into a reward function that closely approximates the true reward function for human driving $r(s_t, a_t)$, then modeling driver behavior reduces to RL.  However, handcrafting an accurate reward function is often difficult, which motivates the use of Generative Adversarial Imitation Learning.

\subsection{Generative Adversarial Imitation Learning}

Although $r(s_{t},a_{t})$ is unknown, a surrogate reward $\tilde{r}(s_{t},a_{t})$ may be learned directly from data, without making use of domain knowledge. GAIL~\cite{ho2016generative} trains a policy to perform expert-like behavior by rewarding it for ``deceiving" a classifier trained to discriminate between policy and expert state-action pairs. Consider a set of simulated state-action pairs $\mathcal{X}_{\theta} =  \left\{(s_1,a_1),(s_2,a_2),...,(s_T,a_T) \right\}$ sampled from $\pi_{\theta}$ and a set of expert pairs $\mathcal{X}_{E}$ sampled from ${\pi_{E}}$.
For a neural network $D_\psi$ parameterized by $\psi$, the GAIL objective is given by:

\begin{equation}
\label{eq:GAIL_obj}
\begin{split}
\underset{\psi}{\max}\ \underset{\theta}{\min}\ V(\theta, \psi) & = \E_{(s,a) \sim \mathcal{X}_{E}}\brock{\log D_{\psi}(s,a)} + \\
             &   \E_{(s,a) \sim \mathcal{X}_{\theta}}\brock{\log (1-D_{\psi}(s,a))}\text{.}
\end{split}
\end{equation}

When fitting $\psi$, \Cref{eq:GAIL_obj} can simply be interpreted as a sigmoid cross entropy objective, maximized by minibatch gradient ascent. Positive examples are sampled from $\mathcal{X}_{E}$ and negative examples are sampled from rollouts generated by  interactions of $\pi_\theta$ with the simulation environment. However, $V(\theta, \psi)$ is non-differentiable with respect to $\theta$, requiring optimization via RL.

In order to fit $\pi_\theta$, a surrogate reward function can be formulated from \cref{eq:GAIL_obj} as:

\begin{equation}
\tilde{r}(s_{t},a_{t}; \psi) = -\log(1-D_{\psi}(s_t, a_t))\text{,}
\end{equation}
which approaches infinity as tuples ($s_t$, $a_t$) drawn from $\mathcal{X}_{\theta}$ become indistinguishable from elements of $\mathcal{X}_{E}$ based on the predictions of $D_\psi$.
After performing rollouts with a given set of policy parameters $\theta$, surrogate rewards $\tilde{r}(s_{t},a_{t};\psi)$ are calculated and TRPO is used to perform a policy update.
 Although $\tilde{r}(s_{t},a_{t};\psi)$ may be quite different from the true reward function optimized by experts, it can be used to drive $\pi_\theta$ into regions of the state-action space similar to those explored by $\pi_{E}$.

\section{DATASET}

We use the public Next-Generation Simulation (NGSIM) datasets for US Highway 101~\cite{systematics2007ngsim} and Interstate 80~\cite{systematics2006ngsim}.
NGSIM provides \num{45} minutes of driving at \SI{10}{Hz} for each roadway.
The US Highway 101 dataset covers an area in Los Angeles approximately \SI{640}{m} in length with five mainline lanes and a sixth auxiliary lane for highway entrance and exit.
The Interstate 80 dataset covers an area in the San Francisco Bay Area approximately \SI{500}{m} in length with six mainline lanes, including a high-occupancy vehicle lane and an onramp.

Traffic density in both datasets transitions from uncongested to full congestion and exhibits a high degree of vehicle interaction as vehicles merge on and off the highway and must navigate in congested flow.
The diversity of driving conditions and the forced interaction of traffic participants makes these sources particularly useful for behavioral studies.
The trajectories were smoothed using an extended Kalman filter~\cite{kalman1960new} on a bicycle model and projected to lanes using centerlines extracted from the NGSIM CAD files.
Cars, trucks, buses, and motorcycles are in the dataset, but only car trajectories were used for model training.

\section{EXPERIMENTS}
In this work, we use GAIL and BC to learn policies for two-dimensional highway driving.
The performance of these policies is subsequently evaluated relative to baseline models.

\subsection{Environment}

All experiments were conducted with the rllab reinforcement learning framework~\cite{duan2016benchmarking}.
The simulation environment is a driving simulation on the NGSIM 80 and 101 road networks.
Simulations are initialized to match frames from the NGSIM data, and the ego vehicle is randomly chosen from among the traffic participants in the frame.
Simulations are run for \num{100} steps at \SI{10}{Hz} and are ended prematurely if the ego vehicle is involved in a collision, drives off road, or drives in reverse.

The ego vehicle is driven according to a bicycle model with acceleration and turn-rate sampled from the policy network.
All other traffic participants are replayed directly from the NGSIM data, but are augmented with emergency braking in the event of an imminent rear-end collision with the ego vehicle.
Specifically, if the acceleration predicted by the Intelligent Driver Model (IDM)~\cite{treiber2000congested} is less than an activation threshold of \SI{-2}{m/s^2}, the vehicle then accelerates according to the IDM while tracking the closest lane centerline.
The IDM is parameterized with a desired speed equal to the vehicle's speed at transition, a minimum spacing of \SI{1}{m}, a desired time headway of \SI{0.5}{s}, a nominal acceleration of \SI{3}{m/s^2}, and a comfortable braking deceleration of \SI{2.5}{m/s^2}.

\subsection{Features}

All experiments use the same set of features.
These features can be decomposed into three sets.
The first set, the \textit{core features}, are eight scalar values that provide basic information about the vehicle's odometry, dimensions, and the lane-relative ego state.
These are listed in \cref{table:core_features}.

\begin{table}[htp]
   \centering
   \caption{\small Core features used by the neural networks.}
   \label{table:core_features}
   \resizebox{0.99\columnwidth}{!}{
     \tiny
     \begin{tabular}{@{}l@{\hskip 5pt}l@{\hskip 5pt}l@{}}
       \toprule
       Feature & Units & Description \\
       \midrule
       Speed & \si{\meter\per\second} &  longitudinal speed \\
       \addlinespace[2pt]
       Vehicle Length & \si{\meter} &  bounding box length \\
       \addlinespace[2pt]
       Vehicle Width & \si{\meter} &  bounding box width \\
       \addlinespace[2pt]
       Lane Offset & \si{\metre} &  lateral centerline offset \\
       \addlinespace[2pt]
       Lane-Relative Heading & \si{\radian} &  heading angle in the Frenet frame \\
       \addlinespace[2pt]
        Lane Curvature & \si{\per\meter} &  curvature of closest centerline point \\
       \addlinespace[2pt]
        Marker Dist. (L) & \si{\meter} &  lat. dist. to left lane marking \\
       \addlinespace[2pt]
        Marker Dist. (R) & \si{\meter} &  lat. dist. to right lane marking \\
       \bottomrule
     \end{tabular}
   }
\end{table}

The core features alone are insufficient to describe the local context.
Information about neighboring vehicles and the local road structure must be incorporated as well.
Several approaches exist for encoding such information in hand-selected features relevant to the driving task~\cite{chen2015deepdriving}.
Rather than restrict the model to a subset of vehicle relationships, we introduce a more general and flexible feature representation.

\begin{figure}[t!]
  \centering
\includegraphics[width=8cm]{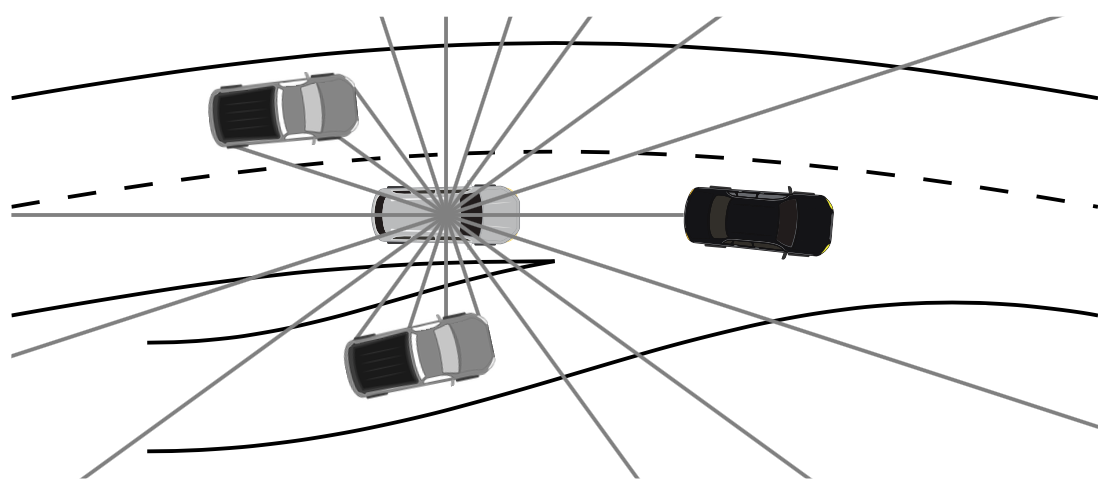}
  \caption{LIDAR-like beams used for measuring range and range rate.}
  \label{fig:lidar}
\end{figure}

In addition to the core features, a set of LIDAR-like beams emanating from the vehicle are used to gather information about its surroundings.
These beams measure both the distance and range rate of the first vehicle struck by them, up to a maximum range.
Our work used a maximum range of \SI{100}{\meter}, with \num{20} range and range rate beams, each spaced uniformly in complete \SI{360}{\degree} coverage around the ego vehicle's center, as shown in \cref{fig:lidar}.

Finally, a set of three indicator features are used to identify when the ego vehicle encounters undesirable states.
These features take on a value of one whenever the ego vehicle is involved in a collision, drives off road, or travels in reverse, and are zero otherwise.
All features were concatenated into a single \num{51}-element vector and fed into each model.

The previous action taken by the ego vehicle is not included in the set of features provided to the policies.
We found that policies can develop an over-reliance on previous actions at the expense of relying on the other features contained in their input.
To counteract this problem, we studied the effect of replacing the previous actions with random noise early in the training process.
However, it was found that even with these mitigations the inclusion of previous actions had a detrimental effect on policy performance.

\subsection{Baseline Models}
The first baseline that we used to compare against our deep policies is a static Gaussian (SG) model, which is an unchanging Gaussian distribution $\pi(a \mid s) = \mathcal{N}(a \mid \mu, \Sigma)$ fit using maximum likelihood.

The second baseline model is a BC approach using mixture regression (MR)~\cite{Lefevre2014}.
The model has been used for model-predictive control and has been shown to work well in simulation and in real-world drive tests.
Our MR model is a Gaussian mixture over the joint space of the actions and features, trained using expectation maximization~\cite{friedman2001elements}.
The stochastic policy is formed from the weighted combination of the Gaussian components conditioned on the features.
Greedy feature selection is used during training to select a subset of predictors up to a maximum feature count threshold while minimizing the Bayesian information criterion~\cite{schwarz1978estimating}.

The final baseline model uses a rule-based controller to govern the lateral and longitudinal motion of the ego vehicle.
The longitudinal motion is controlled by the Intelligent Driver Model with the same parameters as the emergency braking controller used in the simulation environment.
For the lateral motion, MOBIL~\cite{kesting2007general} is used to select the desired lane, with a proportional controller used to track the lane centerline.
A small amount of noise is added to both the lateral and longitudinal accelerations to make the controller nondeterministic.

\subsection{Validation}
To evaluate the relative performance of each model, a systematic validation procedure was performed.
For each model, 1,000 ten-second scenes were simulated 20 times each in an environment identical to the one used to train the GAIL policies.
As these rollouts were performed, several metrics were extracted to quantify the ability of each model to simulate human driver behavior.

\subsubsection{Root-Weighted Square Error}
The root-weighted square error (RWSE) captures the deviation of a model's probability mass from real-world trajectories~\cite{cox2016rwse}.
For predicted variable $v$ over $m$ trajectories, we estimate the RWSE by sampling $n=20$ simulated traces per recorded trajectory:
  \begin{equation}
    \text{RWSE}_H = \sqrt{
    \frac{1}{m n} \sum_{i=1}^m \sum_{j=1}^n \paren{v^{(i)}_H - \hat{v}^{(i,j)}_H}^2
    }\text{,}
  \end{equation}
where $v_H^{(i)}$ is the true value in the $i$th trajectory at time horizon $H$ and $\hat{v}^{(i,j)}_H$ is the simulated variable under sample $j$ for the $i$th trajectory at time horizon $H$.
We extract the RWSE in predictions of global position, centerline offset, and speed over time horizons up to \SI{5}{s}.

\subsubsection{Kullback-Leibler Divergence}
Driver models should produce distributions over emergent quantities that match those observed in real-world data.
For each model, empirical distributions were computed over speed, acceleration, turn-rate, jerk, and inverse time-to-collision (iTTC) over simulated trajectories.
The closeness between the simulated and real-world distributions was quantified using the Kullback-Leibler (KL) divergence.
Piecewise uniform distributions with 100 evenly spaced bins were used.

\subsubsection{Emergent Behavior}
We also extracted a set of emergent metrics that indicate model imitation performance in relation to the NGSIM dataset.
These additional metrics are the lane change rate, the offroad duration, the collision rate, and the hard brake rate.

The lane change rate is the average number of times a vehicle makes a lane change within a 10-second trajectory.
Offroad duration is the average number of time steps per trajectory that a vehicle spends more than \SI{1}{m} outside the closest  outer road marker.
The collision rate is the fraction of trajectories where the ego vehicle intersects with another traffic participant.
The hard brake rate captures the frequency at which a model chooses to brake harder than \SI{-3}{m/s^2}.

The environment in which validation occurs is not entirely realistic, as the non-ego vehicles have pre-recorded trajectories and do not always properly respond to deviations of the ego vehicle from its original trajectory, leading to an artificially high number of collisions.
Hence, we also extract the hard brake rate to help quantify how often dangerous driving situations occur.

\section{RESULTS}

\begin{figure}[t]
  \centering

  \begin{tikzpicture}
    \begin{groupplot}[
      group style={group size= 1 by 3, vertical sep=6pt},
      height=0.4\columnwidth,
      width=\columnwidth,
      enlargelimits=0.0,
      ylabsh=-2.5em,
      ticklabel style = {font=\scriptsize}
    ]

      \nextgroupplot[
        ylabel={\scriptsize Position (\si{m})},
        xticklabels={,,}
      ]

      \begin{texthook}{rwse-pos}
          \addplot[colorA, dotted, thick, mark=none] coordinates{
          	(0,0) (0.500, 0.5441) (1.000, 0.8009) (2.000, 1.6721) (3.000, 2.8785) (4.000, 4.4279) (5.000, 6.1615) };
          \addplot[colorF, densely dotted, thick, mark=none] coordinates{
            (0,0) (0.500, 0.5337) (1.000, 0.6819) (2.000, 1.6343) (3.000, 3.1637) (4.000, 5.0711) (5.000, 7.2548) };
          \addplot[colorG, dashdotted, thick, mark=none] coordinates{
            (0,0) (0.500, 0.5277) (1.000, 0.6379) (2.000, 1.3833) (3.000, 2.5586) (4.000, 3.9919) (5.000, 5.6201) };
          \addplot[colorC, dashed, thick, mark=none] coordinates{
          	(0,0) (0.500, 0.5298) (1.000, 0.6290) (2.000, 1.2713) (3.000, 2.2478) (4.000, 3.3922) (5.000, 4.6609) };
          \addplot[colorE, dashed, thick, mark=none] coordinates{
            (0,0) (0.500, 0.5281) (1.000, 0.6592) (2.000, 1.5744) (3.000, 3.0644) (4.000, 4.9266) (5.000, 7.0623) };
          \addplot[colorC, solid, thick, mark=none] coordinates{
            (0,0) (0.500, 0.5383) (1.000, 0.6888) (2.000, 1.4900) (3.000, 2.6180) (4.000, 3.9105) (5.000, 5.2813) };
          \addplot[colorE, solid, thick, mark=none] coordinates{
            (0,0) (0.500, 0.5337) (1.000, 0.6939) (2.000, 1.5081) (3.000, 2.5528) (4.000, 3.6408) (5.000, 4.7162) };
      \end{texthook}

      \nextgroupplot[
        ylabel={\scriptsize Lane Offset (\si{m})},
        xticklabels={,,}
      ]

      \begin{texthook}{rwse-lane-offset}
          \addplot[colorA, dotted, thick, mark=none] coordinates{
          	(0,0) (0.500, 0.2468) (1.000, 0.4996) (2.000, 0.6303) (3.000, 0.6018) (4.000, 0.5758) (5.000, 0.5845) };
          \addplot[colorF, densely dotted, thick, mark=none] coordinates{
            (0,0) (0.500, 0.0592) (1.000, 0.1418) (2.000, 0.3952) (3.000, 0.6070) (4.000, 0.7617) (5.000, 0.9223) };
          \addplot[colorG, dashdotted, thick, mark=none] coordinates{
            (0,0) (0.500, 0.0589) (1.000, 0.1461) (2.000, 0.3615) (3.000, 0.5569) (4.000, 0.6761) (5.000, 0.7751) };
          \addplot[colorC, dashed, thick, mark=none] coordinates{
          	(0,0) (0.500, 0.0583) (1.000, 0.1381) (2.000, 0.3891) (3.000, 0.5901) (4.000, 0.7055) (5.000, 0.8830) };
          \addplot[colorE, dashed, thick, mark=none] coordinates{
            (0,0) (0.500, 0.0582) (1.000, 0.1377) (2.000, 0.3780) (3.000, 0.5861) (4.000, 0.7165) (5.000, 0.8450) };
          \addplot[colorC, solid, thick, mark=none] coordinates{
            (0,0) (0.500, 0.0866) (1.000, 0.2058) (2.000, 0.4718) (3.000, 0.6472) (4.000, 0.7003) (5.000, 0.7553) };
          \addplot[colorE, solid, thick, mark=none] coordinates{
            (0,0) (0.500, 0.0923) (1.000, 0.2303) (2.000, 0.4897) (3.000, 0.5833) (4.000, 0.6312) (5.000, 0.6837) };
      \end{texthook}

      \nextgroupplot[
        ylabel={\scriptsize Speed (\si{m/s})},
        xlabel=Horizon (\si{s}),
        legend cell align=left,
        legend style={
          draw=none,
          at={(0.5,-0.5)},
          anchor=north,
          legend columns=3,
          font=\scriptsize,
          /tikz/every even column/.append style={column sep=0.5cm}
          }
      ]

      \begin{texthook}{rwse-speed}
          \addplot[colorA, dotted, thick, mark=none] coordinates{
          	(0,0) (0.500, 0.4369) (1.000, 0.7247) (2.000, 1.1834) (3.000, 1.5204) (4.000, 1.7671) (5.000, 1.9771) };
          \addplot[colorF, densely dotted, thick, mark=none] coordinates{
            (0,0) (0.500, 0.4136) (1.000, 0.7712) (2.000, 1.3854) (3.000, 1.8500) (4.000, 2.1939) (5.000, 2.4931) };
          \addplot[colorG, dashdotted, thick, mark=none] coordinates{
          	(0,0) (0.500, 0.3583) (1.000, 0.6530) (2.000, 1.1259) (3.000, 1.4631) (4.000, 1.6946) (5.000, 1.9351) };
          \addplot[colorC, dashed, thick, mark=none] coordinates{
          	(0,0) (0.500, 0.3106) (1.000, 0.5745) (2.000, 0.9654) (3.000, 1.2119) (4.000, 1.3729) (5.000, 1.5448) };
          \addplot[colorE, dashed, thick, mark=none] coordinates{
            (0,0) (0.500, 0.3826) (1.000, 0.7386) (2.000, 1.3472) (3.000, 1.8050) (4.000, 2.1432) (5.000, 2.4419) };
          \addplot[colorC, solid, thick, mark=none] coordinates{
            (0,0) (0.500, 0.3827) (1.000, 0.6821) (2.000, 1.1099) (3.000, 1.3718) (4.000, 1.5247) (5.000, 1.6455) };
          \addplot[colorE, solid, thick, mark=none] coordinates{
            (0,0) (0.500, 0.4357) (1.000, 0.7408) (2.000, 1.1242) (3.000, 1.3069) (4.000, 1.3769) (5.000, 1.4276) };
      \end{texthook}

      \begin{texthook}{model-compare-rwse-legend}
          \legend{IDM+MOBIL, SG, MR, BC MLP, BC GRU, GAIL MLP, GAIL GRU}
        \end{texthook}
      \end{groupplot}
  \end{tikzpicture}

  \caption{\footnotesize
  The root weighted square error for each candidate model vs. prediction horizon. Deep policies outperform the other methods.
  }
  \label{fig:model_compare_rwse}
\end{figure}
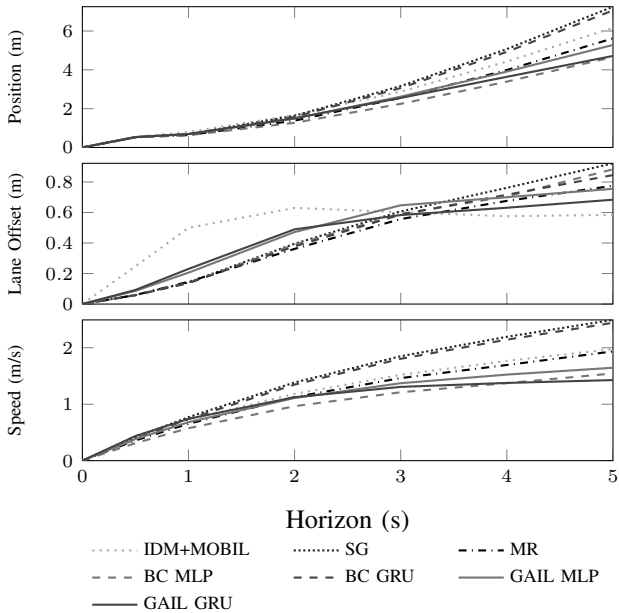

\begin{figure*}[t]
  \centering

  \begin{tikzpicture}
    \begin{texthook}{kldiv}
        \begin{axis}[
          width=1.5\columnwidth,
          height=5cm,
          ylabel={\scriptsize KL divergence},
          ybar=0pt,
          ymin=0.0, ymax=1.5,
          enlarge x limits=0.1,
          legend cell align=left,
          legend style={
            draw=none,
            at={(1.01,0.5)},
            anchor=west,
            legend columns=1,
            font=\scriptsize,
            /tikz/every even column/.append style={column sep=0.5cm}
            },
          bar width=8pt,
          x tick label style={font=\scriptsize}, 
          symbolic x coords={iTTC, Speed, Acceleration, Turn-rate, Jerk},
          xtick=data,
          xtick pos=left,
          ticklabel style = {font=\scriptsize},
        ]
        	\addplot [colorA,fill=colorA!60, postaction={pattern=north west lines, pattern color=white}]
            coordinates{
              (iTTC,        0.0415)
              (Speed,       0.1307)
              (Acceleration,   0.5694)
              (Turn-rate,    0.0628)
              (Jerk,        0.1706)
            };
          \addplot [colorB,fill=colorB!60, postaction={pattern=north east lines, pattern color=white}]
        		coordinates{
        			(iTTC,        0.2560)
        			(Speed,       0.0634)
        			(Acceleration,   0.0457)
        			(Turn-rate,    0.0158)
        			(Jerk,        1.7500)
        		};
        	\addplot [colorC,fill=colorC!60, postaction={pattern=north west lines, pattern color=white}]
        		coordinates{
        			(iTTC,        0.1811)
        			(Speed,       0.0546)
        			(Acceleration,   0.0609)
        			(Turn-rate,    0.0504)
        			(Jerk,        1.6390)
        		};
          \addplot [colorD,fill=colorD!60, postaction={pattern=north east lines, pattern color=white}]
            coordinates{
              (iTTC,        0.1631)
              (Speed,       0.0522)
              (Acceleration,   0.0905)
              (Turn-rate,    0.1278)
              (Jerk,        0.3868)
            };
          \addplot [colorE,fill=colorE!60, postaction={pattern=north west lines, pattern color=white}]
            coordinates{
              (iTTC,        0.2019)
              (Speed,       0.0598)
              (Acceleration,   1.1457)
              (Turn-rate,    0.2482)
              (Jerk,        0.7478)
            };
        	\addplot [colorF,fill=colorF!60, postaction={pattern=north east lines, pattern color=white}]
        		coordinates{
        			(iTTC,        0.1953)
        			(Speed,       0.1002)
        			(Acceleration,   0.1059)
        			(Turn-rate,    0.4736)
        			(Jerk,        0.6010)
        		};
          \addplot [colorG,fill=colorG!60, postaction={pattern=north west lines, pattern color=white}]
            coordinates{
              (iTTC,        0.0788)
              (Speed,       0.0519)
              (Acceleration,   0.0692)
              (Turn-rate,    0.8434)
              (Jerk,        1.0956)
            };
        \legend{IDM+MOBIL, SG, MR, BC MLP, BC GRU, GAIL MLP, GAIL GRU}
        \end{axis}
    \end{texthook}
  \end{tikzpicture}

  \caption{
  \footnotesize
  The KL divergence for various emergent metrics pulled from \SI{10}{s} trajectories.
  }
  \label{fig:KLdiv}
\end{figure*}
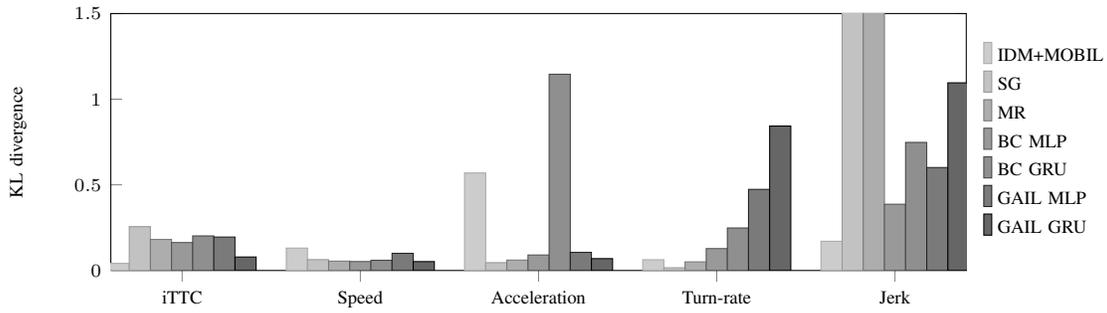

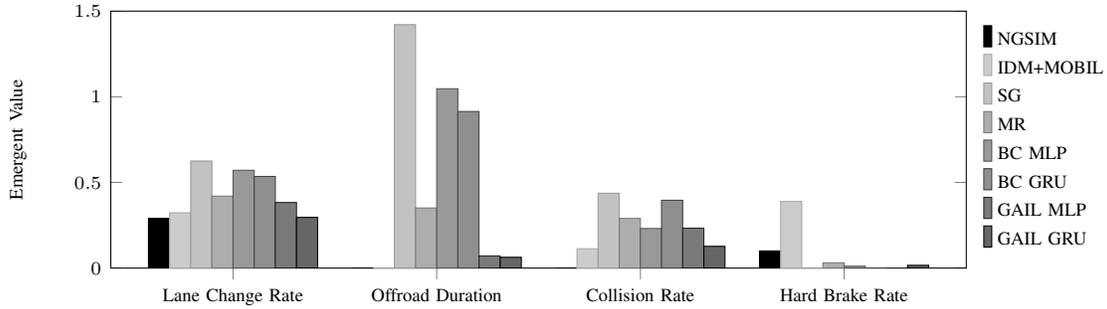
\begin{figure*}[t]
  \centering

  \begin{tikzpicture}
    \begin{texthook}{emergent-values}
        \begin{axis}[
          width=1.5\columnwidth,
          height=5cm,
          ylabel={\scriptsize Emergent Value},
          ybar=0pt,
          ymin=0.0, ymax=1.5,
          enlarge x limits=0.2,
          legend cell align=left,
          legend style={
            draw=none,
            at={(1.01,0.5)},
            anchor=west,
            legend columns=1,
            font=\scriptsize,
            /tikz/every even column/.append style={column sep=0.5cm}
            },
          bar width=8pt,
          x tick label style={font=\scriptsize}, 
          symbolic x coords={Lane Change Rate, Offroad Duration, Collision Rate, Hard Brake Rate},
          xtick=data,
          xtick pos=left,
          ticklabel style = {font=\scriptsize},
        ]

          \addplot [black,fill=black, postaction={pattern=north west lines, pattern color=white}]
            coordinates{
              (Lane Change Rate,   0.290)
              (Offroad Duration,   0.000)
              (Collision Rate,     0.000)
              (Hard Brake Rate,    0.100)
            };
          \addplot [colorA,fill=colorA!60, postaction={pattern=north east lines, pattern color=white}]
            coordinates{
              (Lane Change Rate,   0.323)
              (Offroad Duration,   0.000)
              (Collision Rate,     0.113)
              (Hard Brake Rate,    0.389)
            };
          \addplot [colorB,fill=colorB!60, postaction={pattern=north west lines, pattern color=white}]
            coordinates{
              (Lane Change Rate,   0.625)
              (Offroad Duration,   1.421)
              (Collision Rate,     0.437)
              (Hard Brake Rate,    0.001)
            };
          \addplot [colorC,fill=colorC!60, postaction={pattern=north east lines, pattern color=white}]
            coordinates{
              (Lane Change Rate,   0.420)
              (Offroad Duration,   0.351)
              (Collision Rate,     0.290)
              (Hard Brake Rate,    0.031)
            };
          \addplot [colorD,fill=colorD!60, postaction={pattern=north west lines, pattern color=white}]
            coordinates{
              (Lane Change Rate,   0.571)
              (Offroad Duration,   1.047)
              (Collision Rate,     0.231)
              (Hard Brake Rate,    0.011)
            };
          \addplot [colorE,fill=colorE!60, postaction={pattern=north east lines, pattern color=white}]
            coordinates{
              (Lane Change Rate,   0.535)
              (Offroad Duration,   0.914)
              (Collision Rate,     0.396)
              (Hard Brake Rate,    0.000)
            };
          \addplot [colorF,fill=colorF!60, postaction={pattern=north west lines, pattern color=white}]
            coordinates{
              (Lane Change Rate,   0.383)
              (Offroad Duration,   0.071)
              (Collision Rate,     0.233)
              (Hard Brake Rate,    0.000)
            };
          \addplot [colorG,fill=colorG!60, postaction={pattern=north east lines, pattern color=white}]
            coordinates{
              (Lane Change Rate,   0.296)
              (Offroad Duration,   0.063)
              (Collision Rate,     0.127)
              (Hard Brake Rate,    0.016)
            };
        \legend{NGSIM, IDM+MOBIL, SG, MR, BC MLP, BC GRU, GAIL MLP, GAIL GRU}
        \end{axis}
    \end{texthook}
  \end{tikzpicture}

  \caption{
  \footnotesize
  Emergent values for each model. 
  }
  \label{fig:emergent_values}
\end{figure*}

Validation results for root-weighted square error are given in \cref{fig:model_compare_rwse}.
The root-weighted square error results show that the feedforward BC model has the best short-horizon performance, but then begins to accumulate error for longer time horizons.
GAIL produces more stable trajectories and it short term predictions perform well.
One can clearly see the controller adhere to the lane-centerline, so its lane offset error is close to a constant \num{0.5}, which demonstrates that human drivers do not always closely track the nearest lane-centerline.

KL divergence scores are given in \cref{fig:KLdiv}.
The KL divergence results show very good tracking for SG in everything but jerk.
SG cannot overfit, and always takes the average action. Its performance in other metrics is poor.
GAIL GRU performs well on the iTTC, speed, and acceleration metrics.
It does poorly with respect to turn-rate and jerk.
This poor performance is likely due to the fact that, on average, the GAIL GRU policy takes similar actions to humans, but oscillates between actions more than humans.
For instance, rather than outputting a turn-rate of zero on straight road stretches, it will alternate between outputting small positive and negative turn-rates.
In comparison with the GAIL policies, the BC policies are worse with iTTC.
The GRU version has the largest KL divergence in acceleration, mostly due to its accelerations being generally small in magnitude, but does reasonably well with turn-rate and jerk.

Validation results for emergent variables are given in \cref{fig:emergent_values}.
The emergent values show that GAIL policies outperform the BC policies.
The GAIL GRU policy has the closest match to the data everywhere except for hard brakes (it rarely takes extreme actions).
Mixture regression largely performs better than SG and is on par with the BC policies, but is still susceptible to cascading errors.
Offroad duration is perhaps the most striking statistic; only GAIL (and of course IDM + MOBIL) are able to stay on the road for extended stretches.
SG never brakes hard because it only drives straight, and it has a high collision rate as a consequence.
It is interesting that the collision rate for IDM + MOBIL is roughly the same as the collision rate for GAIL GRU, despite the fact that IDM + MOBIL should not collide.
The inability of other vehicles within the simulation environment to fully react to the ego-vehicle may explain this phenomenon.

The results demonstrate that GAIL-based models capture many desirable properties of both rule-based and machine learning methods, while avoiding common pitfalls.
With the exception of the hand-coded controller, GAIL policies achieve the lowest collision and off-road driving rates, considerably outperforming baseline and similarly structured BC models.
However, GAIL also achieves a lane change rate closer to real human driving than any other method against which it is compared.

Furthermore, extending GAIL to recurrent policies leads to improved performance.
This result is an interesting contrast with the BC policies, where the addition of recurrence largely does not lead to better results.
Thus, we find that recurrence by itself is insufficient for addressing the detrimental effects that cascading errors can have on BC policies.

\section{CONCLUSIONS}

This paper demonstrates the effectiveness of deep imitation learning as a means of training driver models that perform realistically over long time horizons, while simultaneously capturing microscopic, human-like behavior.
Our contributions have been to (1) extend Generative Adversarial Imitation Learning to the optimization of recurrent policies, and to (2)  apply this technique to the creation of a new, intelligent model of highway driving that outperforms the state of the art on several metrics.
Although behavioral cloning still outperforms Generative Adversarial Imitation Learning on short ($\sim$\SI{2}{s}) horizons, its greedy behavior prevents it from achieving realistic driving over an extended period.
The use of \textit{policy optimization} by Generative Adversarial Imitation Learning enables us to overcome this problem of cascading errors to produce long-term, stable trajectories.
Furthermore, the use of \textit{policy representation} by deep, recurrent neural networks enables us to learn directly from general sensor inputs (i.e., LIDAR distance and range rate) that can capture arbitrary traffic states and simulate partial observability.

We have argued that reinforcement learning schemes incorporating surrogate reward functions overcome problems arising from supervised learning in highway driver modeling.
As such, future work may wish to explore ways of combining other reward signals with our own. Whereas Generative Adversarial Imitation Learning captures human-like behavior present in the dataset, simulators may also wish to enforce certain behaviors (e.g., explicitly modeling driver style) by combining the learned, surrogate reward with a reward function crafted from hand-picked features.
An engineered reward function could also be used to penalize the oscillations in acceleration and turn-rate produced by the GAIL GRU.
Finally, we offer our model of human driving behavior as an important element for simulating realistic highway conditions.
Future work will apply our model to decision making and safety validation.
The code associated with this paper can be found at {\tt\small https://github.com/sisl/gail-driver}.




%

\section*{ACKNOWLEDGMENT}

This material is based upon work supported by the Ford Motor Company, Robert Bosch LLC, and the National Science Foundation Graduate Research Fellowship Program under Grant No. DGE-114747.


\printbibliography

\end{document}